# GaMNet: A Hybrid Network with Gabor Fusion and NMamba for Efficient 3D Glioma Segmentation

Chengwei Ye[1][0009-0004-2593-3621], Huanzhen Zhang[2][0009-0008-3051-5642], Yufei Lin[1][0009-0005-0941-3316], Kangsheng Wang*[3][0009-0009-8392-4148], Linuo Xu[4][0009-0004-2901-6193], Shuyan Liu[5][0009-0004-7641-2623]

[1] Homesite Group Inc.
[2] Chewy Inc.
[3] University of Science and Technology Beijing
[4] Yunnan University of Finance and Economics
[5] Yunnan University

\

* Corresponding author.

**Abstract.** Gliomas are aggressive brain tumors that pose serious health risks. Deep learning aids in lesion segmentation, but CNN and Transformer-based models often lack context modeling or demand heavy computation, limiting real-time use on mobile medical devices. We propose GaMNet, integrating the NMamba module for global modeling and a multi-scale CNN for efficient local feature extraction. To improve interpretability and mimic the human visual system, we apply Gabor filters at multiple scales. Our method achieves high segmentation accuracy with fewer parameters and faster computation. Extensive experiments show GaMNet outperforms existing methods, notably reducing false positives and negatives, which enhances the reliability of clinical diagnosis.

## 1 Introduction

### 1.1 A Subsection Sample

Brain tumors are among the deadliest cancers, causing over 250,000 deaths annually [1]. Gliomas, the most common subtype from glial cells [2], have a ~80% two-year mortality rate [3], posing a serious health threat [4]. Accurate diagnosis is essential. Multimodal MRI—including T1, post-contrast T1, T2, and T2-FLAIR—provides detailed tumor information, but manual segmentation is time-consuming and requires radiological expertise [5]. Automated segmentation supports treatment planning by identifying tumor size, location, and morphology.

Despite progress in CNN- and Transformer-based methods [6], glioma segmentation remains difficult due to tumor heterogeneity, shape irregularity, and resolution variation. 3D U-Net [7] improved 3D processing but lacked long-range context. V-Net [8] added residuals; DeepMedic [9] used multi-scale features but had fixed receptive fields. MedNeXt [10] improved data efficiency but remained local. UNETR [11], using Vision Transformers [12], captured global context but required high computation.

Swin UNETR [13] added hierarchical learning via Swin Transformers [14]; BraTS models like ResUNet [15] and Attention U-Net [16] added attention but increased complexity. SegFormer3D [17] used MLP decoders for local-global fusion. nnFormer [18] combined CNNs and self-attention for high accuracy, but with high cost [33].

Mamba [20] efficiently captured long-range dependencies. VMamba [21] used Selective Scan Mechanisms (SSM) for better 2D visual processing, and SegMamba [22] combined SSM encoders with CNN decoders for strong segmentation. However, Mamba-CNN hybrids remain underexplored for glioma.

Many models still struggle to fuse local and global features, causing incomplete segmentation [31]. The enhanced tumor core is small and feature-poor, making detection difficult. Efficient and accurate 3D segmentation remains a key challenge [32].

To address this, we propose a hybrid module that fuses Gabor filters and neural features via the Biological Neural Feature Fusion (BNFF) module. Gabor filters simulate human vision, while multi-scale CNNs and NMamba extract local and global features. The FCSA module captures channel-spatial information, and the QA module highlights lesions across dimensions. Experiments on BraTS2023 and FLARE2022 show our method outperforms state-of-the-art approaches.

## 2      Methods

### 2.1      Overview

In this study, we propose HNGF-NET, based on the U-Net architecture, for improved brain glioma segmentation. The network processes multi-modal medical images, using Gabor filters to simulate the human visual system by extracting spatial and frequency features. The Mixed Feature Extraction (MFE) module captures both fine details and global context, which are fused through the Biological Neural Feature Fusion (BNFF) module. Progressive downsampling enhances edge and texture details. The Quadruple Attention (QA) module refines lesion-specific features along skip connections, while a direct skip from the original feature map improves final-layer upsampling for accurate segmentation.

### 2.2      Gabor Module

Neural networks often struggle to capture multi-scale and multi-orientation features, limiting their ability to represent spatial and frequency characteristics [23]. To address this, we simulate stages of the human visual system—retinal ganglion cells (RGC), lateral geniculate nucleus (LGN), and primary visual cortex (V1)—using Gabor filters [24]. These filters extract features across scales and orientations in both spatial and frequency domains. Spatially, they are robust to translation, rotation, and scaling; in the frequency domain, they effectively distinguish textures by responding to local frequency variations [25]. This enhances feature richness and improves biological interpretability [26].

The Gabor function combines a Gaussian and a cosine component, with orthogonal real and imaginary parts. The real part smooths images, while the imaginary part emphasizes edges. The 3D Gabor filter is defined as follows:

$$G(x,y,z;\lambda,\theta,\phi,\Psi,\sigma,\gamma_1,\gamma_2) = \exp\left(-\frac{x'^2 + \gamma_1^2 y'^2 + \gamma_2^2 z'^2}{2\sigma^2}\right) \exp\left(i\left(2\pi\frac{x'}{\lambda} + \Psi\right)\right) \quad (1)$$

Real part:

$$G(x,y,z;\lambda,\theta,\phi,\Psi,\sigma,\gamma_1,\gamma_2) = \exp\left(-\frac{x'^2 + \gamma_1^2 y'^2 + \gamma_2^2 z'^2}{2\sigma^2}\right) \cos\left(2\pi\frac{x'}{\lambda} + \Psi\right) \quad (2)$$

Imaginary part:

$$G(x,y,z;\lambda,\theta,\phi,\Psi,\sigma,\gamma_1,\gamma_2) = \exp\left(-\frac{x'^2 + \gamma_1^2 y'^2 + \gamma_2^2 z'^2}{2\sigma^2}\right) \sin\left(2\pi\frac{x'}{\lambda} + \Psi\right) \quad (3)$$

x′, y′ and z′:

$$\begin{pmatrix} x' \\ y' \\ z' \end{pmatrix} = \begin{pmatrix} \sin\phi & 0 & \cos\phi \\ 0 & 1 & 0 \\ -\cos\phi & 0 & \sin\phi \end{pmatrix} \times \begin{pmatrix} \cos\theta & \sin\theta & 0 \\ -\sin\theta & \cos\theta & 0 \\ 0 & 0 & 1 \end{pmatrix} \times \begin{pmatrix} x \\ y \\ z \end{pmatrix} \quad (4)$$

where x′, y′, and z′ are coordinates adjusted relative to the center position in space. λ is the wavelength of the sinusoidal plane wave, θ is the angle between the Gabor filter and the x-axis, φ is the angle relative to the z-axis, σ represents the standard deviation of the Gaussian kernel, and γ is the spatial aspect ratio.

### 2.2 Mixed Feature Extraction (MFE) Module

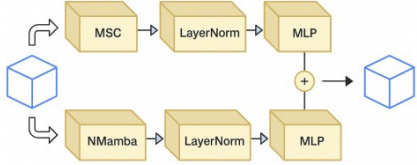

**Fig. 1.** MFE module

In brain MRI, gliomas often resemble healthy tissue in texture, causing misclassification and reduced accuracy [27]. To address this, we introduce a Mixed Feature Extraction (MFE) module that captures both local and global features [28]. As shown in Fig. 1, the input passes through the MSC module with 5×5×5, 3×3×3, and 1×1×1 convolutions to extract scale-specific local features, while the NMamba module extracts global features with fewer parameters and lower complexity [29]. Both outputs undergo Layer Normalization and MLP processing, then are fused via addition to integrate multi-scale and contextual information [30].

In the following sections, we will provide a detailed introduction to these components.

**MSC Module.** Traditional convolution layers often struggle to capture the diverse features in brain MRI, especially across varying sizes and complexities [21]. To address this, we introduce a cascade residual Multi-Scale Convolution (MSC) module using 5×5×5, 3×3×3, and 1×1×1 kernels. Each kernel targets a specific spatial scale: 1×1×1 captures fine details, 3×3×3 focuses on mid-level features, and 5×5×5 provides broader contextual information [25]. As shown in Fig. 2, this design enables deep local feature extraction. To further enhance the model's comprehension and generalization, we incorporate the FCSA module [28].

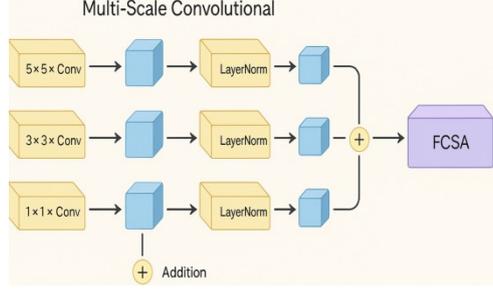

**Fig. 2.** MSC module

This method of cascade-based multi-scale residual comprehensively captures the local details of the image, thereby enhancing the accuracy and stability of glioma segmentation. Input a 3D data $X \in \mathbb{R}^{C \times H \times W \times D}$, which is computed as:

$$\begin{aligned}
X^1 &= RELU(LN(Conv\ 5 \times 5 \times 5(X))) \\
X^2 &= RELU(LN(Conv\ 3 \times 3 \times 3(X))) \\
X^3 &= RELU(LN(Conv\ 1 \times 1 \times 1(X))) \\
X^4 &= RELU(LN(Conv\ 1 \times 1 \times 1(X^2 + X^3))) \\
X^5 &= RELU(LN(Conv\ 1 \times 1 \times 1(X^1 + X^3))) \\
X^6 &= RELU(LN(Conv\ 1 \times 1 \times 1(X^1 + X^2))) \\
Y &= FCSA(X^4 + X^5 + X^6)
\end{aligned} \quad (5)$$

where Conv(·) denotes the 3D convolutional layer, LN(·) denotes the Layer Normalization and FCSA is the Feature Channel spatial Attention module.

**NMamba Module.** Traditional Transformers are computationally expensive for 3D medical images due to high complexity [15]. Mamba addresses this with a structured state-space model achieving linear complexity (1)(2)(3)(4)(5), reducing computational load. However, larger images make it harder to retain early-stage information. To address this, we propose NMamba, which integrates the Feature Channel and Spatial Attention (FCSA) module into Mamba. FCSA removes noise and preserves critical channel and spatial features [8], enhancing representation and reducing memory loss for efficient large-scale processing.

As illustrated in Fig. 3, the Mamba block reshapes the 3D feature data into a 1D long sequence, which minimizes perceptual bias and directs the most specific attention to the features rather than the spatial structure. After the Mamba block, they are subsequently restored to their 3D shape by reshaping. Input a 3D data $X \in \mathbb{R}^{C \times H \times W \times D}$, which is computed as:

$$X^m = Reshape(MLP(Mamba(Flatten(X))))$$
$$Z = FCSA(Reshape(MLP(Mamba(Flatten(X^m)))))$$
(6)

where Flatten (·) denotes convert data from 3D to 1D, Mamba (·) denotes Mamba block and MLP (·) denotes multi-layer perceptron, Reshape (·) denotes convert data from 1D to 3D and FCSA (·) is the Feature Channel spatial Attention module.

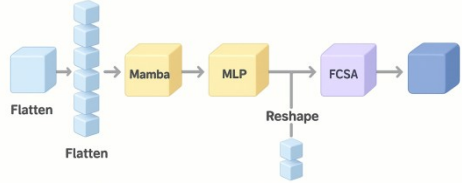

**Fig. 3.** NMamba module

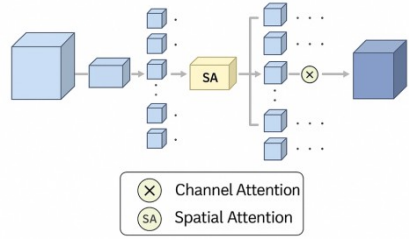

**Fig. 4.** FCSA module

**FCSA Module.** General convolutional networks often struggle to suppress background noise and emphasize relevant features, hindering accurate tumor boundary delineation [12]. The Feature Channel Spatial Attention (FCSA) module addresses this by applying channel and spatial attention to enhance feature fusion [26], improving discrimination and target localization. It highlights local details, emphasizes tumor regions, and suppresses irrelevant information, boosting segmentation performance [15].

As illustrated in Fig. 4, to more efficiently utilize this feature information, we first divide the feature maps into N equal-sized feature vectors along the height (H), width (W) and depth (D) dimensions. For the divided feature vectors $Y_{i,j,k}$ (where i=1…H, j=1…W, k=1…D), we initially introduce a Channel Attention (CA) mechanism. The CA mechanism emphasizes important channel information in the feature vectors through weight allocation $\alpha_{i,j,k}$ (where i=1…H, j=1…W, k=1…D), thereby generating channel attention feature maps , , $Y^c_{i,j,k}$ (where i=1…H, j=1…W, k=1…D). By suppressing redundant information, this operation enhances feature maps by strengthening critical tumor-related channel features. We then apply a Spatial

Attention (SA) mechanism to identify key tumor regions, including boundaries and texture-dense areas. It detects significant spatial regions and allocates them higher weights $\beta_{i,j,k}$ (where i=1…H, j=1…W, k=1…D), thereby generating spatial attention feature maps $Y^s_{i,j,k}$ (where i=1…H, j=1…W, k=1…D). Input a 3D data $Y \in \mathbb{R}^{C \times H \times W \times D}$, which are computed as follows:

$$CA(Y) = Sigmoid(MLP(AvgPool(Y)) + MLP(MaxPool(Y)))$$
$$SA(Y) = Sigmoid(Conv([AvgPool(Y); MaxPool(Y)]))$$
$$Y = \{Y_{i,j,k} | i=1…H; j=1…W; k=1…D\}$$
$$Y^c_{i,j,k} = CA(Y_{i,j,k}) \times \alpha_{i,j,k} \quad (7)$$
$$Y^s_{i,j,k} = SA(Y^c_{i,j,k}) \times \beta_{i,j,k}$$
$$F = \sum_{i=1}^{H}\sum_{j=1}^{W}\sum_{k=1}^{D} Reconstruct(Y^s_{i,j,k})$$

where Conv(·) denotes the 3D convolutional layer, MLP(·) denotes the multi-layer perceptron, MaxPool(·) denotes the max pooling and AvgPool(·) denotes the average pooling, $\alpha_{i,j,k}$ and $\beta_{i,j,k}$ denote the weight factors and Reconstruct(·) denotes the location where the subblock is mapped back to the original feature map.

By introducing this multi-dimensional attention mechanism, the model can capture more comprehensive feature representations, effectively enhancing its ability to characterize the tumor region. Finally, all attention feature maps $Y^c_{i,j,k}$ (where i=1…H, j=1…W, k=1…D) and $Y^s_{i,j,k}$ (where i=1…H, j=1…W, k=1…D) are reintegrated to generate the comprehensive attention feature map F for the current sampling stage.

### 2.3 Biological Neural Feature Fusion (BNFF) Module

**BNFF Module.** Traditional models often struggle to integrate multi-source features, resulting in incomplete tumor representations. The Biological Neural Feature Fusion (BNFF) module fuses features from Gabor filters and the Mixed Feature Extraction (MFE) module. As shown in Fig. 5, it uses two parallel branches for global aggregation and squeeze-and-excitation. In the i-th BNFF, global average pooling captures channel-wise information, followed by two convolution layers with ReLU and Sigmoid activations to adjust weights. Outputs are fused via element-wise addition (6)(7), enhancing feature diversity, robustness, and downstream discrimination.

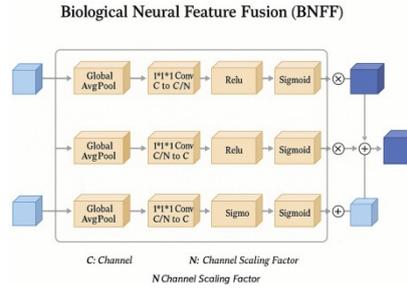

**Fig. 5.** BNFF module

## 2.4 Quadruple Attention (QA) Module

Traditional attention mechanisms often overlook spatial-channel interactions due to limited dimensional focus. To better process 3D data, we extend triplet attention to four dimensions: channel (C), height (H), width (W), and depth (D). As shown in Fig. 6, the Quadruple Attention module generates four branches—(H, C, W, D), (W, C, H, D), (D, C, H, W), and (C, H, W, D)—via tensor permutation. The first three associate channel with spatial dimensions, while the last highlights spatial features. Averaging all branches reduces reliance on any one and improves robustness.

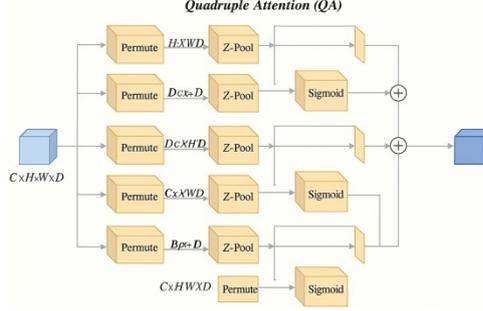

**Fig. 6.** Quadruple Attention module

Input a 3D data $M \in \mathbb{R}^{C \times H \times W \times D}$, which are computed as follows:

$$M^H = Permute^{H \times C \times W \times D}(M)$$
$$M^W = Permute^{W \times C \times H \times D}(M) \quad (8)$$
$$M^D = Permute^{D \times C \times H \times W}(M)$$

where Permute(·) denotes transpose the dimensions of the tensor function.

$$Z - pool(M) = Cat(MaxPool(M), AvgPool(M)) \quad (9)$$

where Cat(·) is the dimension concatenation function, MaxPool(·) and AvgPool(·) denote that the max pooling and average pooling occur in the channel dimension, the spatial-wise branch is represented as:

$$QA^C(M) = X \times Sigmoid(Conv(Z - Pool(M)))$$
$$QA^H(M^H) = Permute^{C \times H \times W \times D}(M^H \times Sigmoid(Conv(Z - Pool(M^H))))$$
$$QA^W(M^W) = Permute^{C \times H \times W \times D}(M^W \times Sigmoid(Conv(Z - Pool(M^W)))) \quad (10)$$
$$QA^D(M^D) = Permute^{C \times H \times W \times D}(M^D \times Sigmoid(Conv(Z - Pool(M^D))))$$

where Sigmoid(·) is the Sigmoid function and Conv(·) denotes the 3D convolutional layer.

Lastly, the output of the Quadruple Attention module is represented as:

$$QA^{out}(M) = \frac{1}{4}\left(QA^H(M^H) + QA^W(M^W) + QA^D(M^D) + QA^C(M)\right) \quad (11)$$

## 3 Experiments

### 3.1 Datasets and Metrics

We conducted experiments on the BraTS2023 and FLARE2022 datasets. BraTS2023 was expert-preprocessed (alignment, skull stripping, interpolation) for consistent resolution, with each volume containing 155 2D slices and annotated tumor subregions: peritumoral edema (ED), enhanced tumor (ET), and necrotic core (NCR). FLARE2022 provides 50 annotated samples across 13 abdominal organs, including the liver, spleen, pancreas, kidneys, stomach, and others. We evaluated performance using Dice Score (Dice), Hausdorff Distance 95% (HD95), False Negative Rate (FNR), and False Positive Rate (FPR), enabling comprehensive comparison in brain glioma segmentation.

### 3.2 Implementation Details

All experiments were conducted on a single NVIDIA GeForce RTX3090 GPU. The dataset was split into training, validation, and test sets in a 7:1:2 ratio. Training ran for 1500 epochs with validation every 20 epochs. To enhance generalization, we applied online data augmentation, including random rotation, cropping, scaling, and mirror flipping. We used cross entropy loss and optimized with SGD and a 'poly' learning rate scheduler (learning rate 1e-3, momentum 0.99, decay 1e-5). No pretraining weights were used to ensure fair evaluation.

**Table 1.** QUANTITATIVE RESULTS FOR DIFFERENT METHODS INTERMS OF DICE AND HD95 ON BRATS2023 DATASET

| Methods | Year | Type | Dice(%)↑ | | | | HD95↓ | | | |
|---|---|---|---|---|---|---|---|---|---|---|
| | | | WT | TC | ET | Avg | WT | TC | ET | Avg |
| UNETR[11] | 2022 | MICCAI | 90.75 | 86.39 | 80.96 | 86.03 | 6.69 | 5.77 | 6.38 | 6.28 |
| Swin UNETR[13] | 2022 | CVPR | **92.97** | 88.86 | 81.90 | 87.91 | 4.09 | 4.36 | 5.25 | 4.57 |
| nnFormer[18] | 2023 | IEEE TIP | 92.39 | 88.52 | 82.30 | 87.74 | 4.34 | 4.36 | 4.99 | 4.56 |
| MedNeXt[10] | 2023 | MICCAI | 92.52 | 88.11 | 81.13 | 87.25 | 4.83 | 4.24 | 5.27 | 4.78 |
| UNETR++[19] | 2024 | IEEE TMI | 91.84 | 84.83 | 80.88 | 85.85 | 4.37 | 5.62 | 5.44 | 5.14 |
| SegFormer3D[17] | 2024 | CVPR Workshop | 89.64 | 83.87 | 73.88 | 82.46 | 6.41 | 6.11 | 6.99 | 6.50 |
| SegMamba[22] | 2024 | MICCAI | 92.85 | 88.87 | 81.93 | 87.88 | **3.95** | 4.19 | 5.11 | 4.42 |
| **Ours** | 2025 | - | 92.69 | **89.02** | **83.35** | **88.35** | 4.05 | **4.15** | **4.75** | **4.32** |

**Table 2.** QUANTITATIVE RESULTS FOR DIFFERENT METHODS INTERMS OF FNR AND FPR ON BRATS2023 DATASET

| Methods | Year | Type | FNR↓ | | | | FPR↓ | | | |
|---|---|---|---|---|---|---|---|---|---|---|
| | | | WT | TC | ET | Avg | WT | TC | ET | Avg |
| UNETR[11] | 2022 | MICCAI | 0.06 | 0.08 | 0.15 | 0.10 | 0.0012 | 0.0006 | 0.03 | 0.011 |
| Swin UNETR[13] | 2022 | CVPR | 0.06 | 0.08 | 0.18 | 0.11 | **0.0007** | 0.0045 | 0.04 | 0.015 |
| nnFormer[18] | 2023 | IEEE TIP | 0.06 | 0.07 | 0.15 | 0.09 | 0.0008 | 0.0165 | 0.04 | 0.019 |
| MedNeXt[10] | 2023 | MICCAI | 0.05 | **0.05** | 0.13 | 0.08 | 0.0008 | 0.0006 | 0.04 | 0.014 |
| UNETR++[19] | 2024 | IEEE TMI | 0.04 | 0.07 | 0.14 | 0.08 | 0.0011 | 0.0247 | 0.04 | 0.022 |
| SegFormer3D[17] | 2024 | CVPR Workshop | 0.05 | 0.13 | 0.27 | 0.15 | 0.0012 | 0.0045 | 0.03 | 0.012 |
| SegMamba[22] | 2024 | MICCAI | 0.05 | 0.08 | 0.12 | 0.08 | 0.0008 | 0.0044 | 0.04 | 0.015 |
| **Ours** | 2025 | - | **0.03** | 0.07 | **0.11** | **0.07** | 0.0009 | **0.0005** | **0.03** | **0.010** |

**Table 3.** QUANTITATIVE RESULTS FOR DIFFERENT METHODS INTERMS OF DICE ON FLARE2022 DATASET

| Methods | Spl↑ | Rkid↑ | Lkid↑ | Gall↑ | Eso↑ | Liv↑ | Sto↑ | Aor↑ | Ivc↑ | Duo↑ | Pan↑ | Rag↑ | Lag↑ | Avg↑ |
|---|---|---|---|---|---|---|---|---|---|---|---|---|---|---|
| UNETR[11] | 95.38 | 92.87 | 94.51 | 75.18 | 91.06 | 83.27 | 79.65 | 80.35 | 79.98 | 65.25 | 83.57 | 64.18 | 92.74 | 82.92 |
| Swin UNETR[13] | 98.15 | 94.09 | 97.44 | 82.11 | 95.34 | 90.81 | 84.59 | 84.81 | 86.77 | 80.97 | 94.27 | 79.63 | 96.50 | 89.65 |
| nnFormer[18] | 97.94 | 94.27 | 96.22 | 84.61 | 95.30 | 90.58 | 83.27 | 85.04 | 89.41 | **85.29** | 95.38 | 81.89 | 95.53 | 90.36 |
| MedNeXt[10] | **98.71** | 93.68 | 96.38 | **85.20** | 95.60 | 91.14 | 84.61 | 85.02 | **90.63** | 82.05 | **95.67** | 84.81 | 95.37 | 90.68 |
| UNETR++[19] | 98.25 | 94.27 | 97.59 | 82.74 | 95.37 | 91.25 | 82.71 | 82.80 | 87.87 | 76.78 | 95.25 | 77.62 | 96.88 | 89.18 |
| SegFormer3D[17] | 96.49 | 92.42 | 95.86 | 72.94 | 91.60 | 84.00 | 68.24 | 68.23 | 80.64 | 69.08 | 89.83 | 68.28 | 94.01 | 82.43 |
| SegMamba[22] | 98.30 | 93.24 | **97.89** | 83.20 | 95.91 | 91.25 | **85.31** | 85.63 | 89.39 | 83.64 | 94.30 | 82.08 | 96.96 | 90.55 |
| **Ours** | 97.93 | **94.31** | 96.46 | 83.15 | **95.97** | **91.30** | 84.03 | **85.80** | 89.30 | 84.57 | 94.47 | **84.87** | **96.98** | **90.70** |

**Table 4.** QUANTITATIVE RESULTS FOR DIFFERENT METHODS INTERMS OF HD95 ON FLARE2022 DATASET

| Methods | Spl↓ | Rkid↓ | Lkid↓ | Gall↓ | Eso↓ | Liv↓ | Sto↓ | Aor↓ | Ivc↓ | Duo↓ | Pan↓ | Rag↓ | Lag↓ | Avg↓ |
|---|---|---|---|---|---|---|---|---|---|---|---|---|---|---|
| UNETR[11] | 47.92 | 28.09 | 42.44 | 72.55 | 6.41 | 10.82 | 2.95 | 12.87 | 8.43 | 77.18 | 122.63 | 132.66 | 38.9 | 46.45 |
| Swin UNETR[13] | 2.36 | 6.22 | 14.24 | 12.55 | 1.74 | 4.73 | 9.21 | 2.25 | 4.33 | 38.05 | 46.03 | 12.75 | 12.4 | |
| nnFormer[18] | 4.85 | 14.91 | 20.46 | 5.68 | 2.7 | 5.79 | 2.12 | 1.93 | 3.76 | 5.29 | 3.21 | 11.4 | 2.65 | 6.52 |
| MedNeXt[10] | **1.69** | 4.45 | 3.5 | 5.18 | 4.75 | 4.08 | **1.78** | 3.87 | 3.19 | 17.52 | 8.36 | 4.84 | 5.03 | |
| UNETR++[19] | 1.97 | 4.26 | **2.13** | 34.27 | 2.43 | 4.59 | 1.95 | 26.93 | 4.93 | 106.29 | 3.3 | 147.26 | 2.29 | 26.35 |
| SegFormer3D[17] | 8.55 | 8.33 | 3.52 | 10.38 | 2.92 | 6.99 | 3.87 | 4.29 | 8.58 | 11.06 | 8.26 | 14.04 | 4.54 | 7.33 |
| SegMamba[22] | 6.03 | 4.85 | 2.26 | 9.28 | **1.55** | 4.61 | 1.81 | 1.93 | 3.94 | 16.9 | 33.86 | 8.47 | 2.27 | 7.52 |
| **Ours** | 8.39 | **4.04** | 3.47 | 7.47 | 1.71 | **4.03** | 7.92 | **1.41** | 5.39 | **4.29** | 6.98 | **6.49** | **2.26** | **4.91** |

**Table 5.** QUANTITATIVE RESULTS FOR DIFFERENT METHODS INTERMS OF FNR ON FLARE2022 DATASET.(ALL DATA HAVE BEEN MULTIPLIED BY 100)

| Methods | Spl↓ | Rkid↓ | Lkid↓ | Gall↓ | Eso↓ | Liv↓ | Sto↓ | Aor↓ | Ivc↓ | Duo↓ | Pan↓ | Rag↓ | Lag↓ | Avg↓ |
|---|---|---|---|---|---|---|---|---|---|---|---|---|---|---|
| UNETR[11] | 0.400 | 0.035 | 0.035 | 0.086 | 0.021 | 0.026 | 0.001 | 0.002 | 0.010 | 0.007 | 0.263 | 0.061 | 0.055 | 0.077 |
| Swin UNETR[13] | 0.118 | 0.030 | 0.020 | 0.062 | 0.011 | 0.024 | 0.002 | 0.002 | 0.009 | 0.007 | 0.069 | 0.036 | 0.022 | 0.032 |
| nnFormer[18] | 0.125 | 0.032 | 0.032 | **0.035** | 0.015 | 0.030 | 0.002 | 0.003 | 0.010 | 0.006 | 0.062 | 0.034 | 0.020 | 0.031 |
| MedNeXt[10] | **0.086** | 0.026 | 0.013 | 0.045 | 0.019 | 0.018 | 0.001 | 0.002 | 0.006 | 0.004 | **0.059** | 0.032 | **0.016** | 0.025 |
| UNETR++[19] | 0.092 | 0.027 | 0.020 | 0.053 | 0.010 | 0.020 | 0.002 | 0.003 | 0.011 | 0.011 | 0.061 | 0.056 | 0.018 | 0.030 |
| SegFormer3D[17] | 0.217 | 0.031 | 0.034 | 0.039 | 0.012 | 0.041 | 0.004 | 0.004 | 0.012 | 0.006 | 0.097 | 0.055 | 0.022 | 0.044 |
| SegMamba[22] | 0.102 | 0.022 | 0.013 | 0.050 | 0.009 | 0.018 | **0.001** | 0.002 | 0.009 | 0.006 | 0.060 | 0.031 | 0.017 | 0.026 |
| **Ours** | 0.099 | **0.017** | **0.013** | 0.041 | **0.008** | **0.012** | 0.003 | **0.001** | **0.006** | **0.004** | 0.070 | **0.023** | 0.017 | **0.024** |

**Table 6.** QUANTITATIVE RESULTS FOR DIFFERENT METHODS INTERMS OF FPR ON FLARE2022 DATASET

| Methods | Spl↓ | Rkid↓ | Lkid↓ | Gall↓ | Eso↓ | Liv↓ | Sto↓ | Aor↓ | Ivc↓ | Duo↓ | Pan↓ | Rag↓ | Lag↓ | Avg↓ |
|---|---|---|---|---|---|---|---|---|---|---|---|---|---|---|
| UNETR[11] | 0.02 | 0.08 | 0.06 | 0.16 | 0.10 | 0.21 | 0.27 | 0.24 | 0.23 | 0.42 | 0.12 | 0.39 | 0.05 | 0.18 |
| Swin UNETR[13] | 0.01 | 0.06 | **0.02** | 0.12 | 0.05 | 0.10 | 0.17 | 0.17 | 0.15 | 0.22 | 0.06 | 0.23 | 0.03 | 0.11 |
| nnFormer[18] | 0.02 | 0.06 | 0.04 | 0.14 | **0.04** | 0.08 | 0.18 | 0.14 | **0.10** | 0.18 | 0.05 | 0.21 | 0.05 | 0.10 |
| MedNeXt[10] | 0.02 | 0.06 | 0.02 | 0.13 | 0.06 | 0.10 | 0.20 | 0.15 | 0.11 | 0.21 | 0.05 | 0.20 | 0.03 | 0.10 |
| UNETR++[19] | 0.02 | 0.06 | 0.02 | 0.12 | 0.06 | 0.10 | 0.17 | 0.14 | 0.11 | 0.22 | 0.04 | 0.19 | 0.03 | 0.10 |
| SegFormer3D[17] | 0.03 | 0.09 | 0.04 | 0.31 | 0.12 | 0.17 | 0.32 | 0.35 | 0.23 | 0.40 | 0.11 | 0.35 | 0.08 | 0.20 |
| SegMamba[22] | 0.01 | **0.05** | 0.03 | 0.12 | 0.05 | 0.10 | 0.19 | 0.15 | 0.11 | **0.18** | 0.05 | 0.22 | 0.03 | 0.10 |
| **Ours** | **0.01** | 0.06 | 0.04 | **0.12** | 0.05 | **0.08** | **0.17** | **0.11** | 0.12 | 0.24 | **0.03** | **0.18** | **0.02** | **0.09** |

### 3.3 Quantitative Comparisons

We evaluated our method against several advanced models. The Wilcoxon signed-rank test confirmed statistically significant improvements ($p<0.05$). As shown in Table 1 and Table 2, our model outperforms others on Dice (88.35%), HD95 (4.32), FNR (0.07), and FPR (0.01). It achieves the best Dice and FNR in the enhancing tumor (ET) region and matches SegMamba [22] on HD95 for the whole tumor (WT). For FPR in the tumor core (TC), it yields the lowest value, reflecting precise segmentation and fewer diagnostic errors.

On the FLARE2022 dataset (Table 3–6), our model achieves an average Dice of 90.70 and HD95 of 4.91, with the best FNR and FPR, especially for the right and left kidneys, reducing missed detections and enhancing clinical reliability.

Figure 8 shows that other models often under- or over-segment lesions, either missing tumor areas or misclassifying normal tissue, complicating clinical decisions.

Overall, our model consistently delivers superior quantitative and visual results, enabling accurate, reliable glioma segmentation with clear boundaries and minimal misclassification.

### 3.4   Ablation Study

To ensure reliable ablation comparisons, we modify only one component at a time while keeping others fixed. As shown in Table VII, removing Gabor filters slightly reduces performance (Dice: 88.18, HD95: 4.41). Excluding the Multi Scale Convolution (MSC) causes a larger drop (Dice: 87.25, HD95: 4.83). Removing NMamba yields a Dice of 87.41, FSCA 87.82, BNFF 87.66, and QA 88.13. These results highlight the contribution of each component to improved segmentation, with reduced false negatives and false positives.

### 3.5   Efficiency Study

To assess our network's balance between efficiency and accuracy, we use Flops, Param, Dice score, and HD95. As shown in Table VIII, our model offers notable advantages. With only 47.4M parameters, it is the most efficient, smaller than nnFormer [18], SegMamba [22], MedNeXt [10], and Swin UNETR [13]. Its 2539.5G Flops is higher than MedNeXt [10] and nnFormer [18], but lower than SegMamba [22]. It achieves the highest Dice (88.35), and the lowest HD95 (4.32), FNR (0.07), and FPR (0.010), ensuring precise segmentation and fewer false detections.

In summary, our architecture offers low complexity and strong segmentation performance.

## 4   Conclusion

We propose a mixed feature extraction module that integrates Gabor filters—simulating the human visual system—into the neural network for multi-level feature extraction. MSC and NMamba encoders capture fine-grained and global tumor features, while FCSA enhances channel-spatial extraction, and QA highlights lesion regions across dimensions. Experiments on BraTS2023 and FLARE2022 confirm our method outperforms state-of-the-art approaches.

Future work will explore integrating MRI with other modalities (e.g., PET, CT) and non-imaging data (e.g., clinical reports, genomics) to further improve accuracy, generalization, and clinical applicability in brain glioma diagnosis and treatment.